# Group theory, group actions, evolutionary algorithms, and global optimization

Andrew Clark[a]

Lipper, a Thomson Reuters company, 707 17th Street, 22nd Floor, Denver, CO 80202, USA

**A B S T R A C T**

In this paper we explain, using dynamical systems analysis, how evolutionary algorithms (EAs) solve global optimization problems. We start with a basic definition of EAs and a baseline random heuristic. Vose and Wright's 1998 generalization of the Vose simple evolutionary algorithm model allows us to characterize alphabets of different lengths as multiplication and addition over rings of integers. We use the related group actions to introduce orbits, equivalence relations, and invariants to the operation of EAs.

## 1. Introduction

This work discusses the efforts of several authors to explain how EAs work. The emphasis of this paper is the use of group theory and in particular group actions. From this baseline a simple connection is made to dynamical systems, in particular specific gradient systems that show how EAs work.

### 1.1 Brief introduction to EAs

A simple EA (also known as a genetic algorithm [GA]) assumes a discrete search space $H$ and a function $f : H \to \mathbb{R}$. The general problem is to find $\arg\min_{x \in H} f$, where $x$ is a vector of the decision variables and $f$ is the objective function.

With EAs it is customary to distinguish genotype–the encoded representation of the variables–from phenotype–the set of variables themselves. The vector $x$ is represented by a string *s* of length *l* made up of symbols drawn from an alphabet *A* using the mapping $c: A^l \to H$. The string length *l* depends on the dimensions of both *H* and *A*, with the elements of the string corresponding to genes and the values to alleles. This statement of genes and alleles is often referred to as genotype-phenotype mapping.

---
[a] Tel: +13039416017
 E-mail address: andrew.clark@thomsonreuters.com

With EAs it is helpful if *c* is a bijection. The important property of bijections as they apply to EAs is that bijections have an inverse, i.e., there is a unique vector *x* for each string *s* and a unique string *s* for each vector *x*.

The execution of an EA typically begins by randomly sampling with replacement from $A^l$. The resulting collection is the initial population, denoted by $P_0$. In general, a population is a collection $P = <s_1, s_2, ..., s_\mu>$ of individuals, where $s_i \in H$, and populations are treated as n-tuples of individuals (as signified by the angle brackets). The number of individuals *μ* is defined as the population size.

Following initialization, execution proceeds iteratively. Each iteration consists of an application of one or more of the evolutionary operators (EOs): crossover, mutation, and selection. The combined effect of the EOs applied in a particular generation $t \in \mathbb{N}$ transforms the current population *P(t)* into a new population *P(t+1)*.

In the population transformation $\mu, \mu' \in \mathbb{Z}^+$ (the parent and offspring population sizes, respectively) the mapping $\Upsilon: H^\mu \to H^{\mu'}$ is called a population transform (PT). If $\Upsilon(P) = P'$, then *P* is a parent population and $P'$ is the offspring population. If $\mu = \mu'$, it is called the population size.

The PT resulting from an EO often depends on the outcome of a random experiment. In Merkle and Lamont [1] this result is referred to as a random population transformation (RPT or random PT). To define an RPT, let $\mu \in \mathbb{Z}^+$ and $\Omega$ be a set—a sample space. A random function

$R: \Omega \to \Upsilon(H^\mu, \bigcup_{\mu' \in \mathbb{Z}^+} H^{\mu'})$ is called an RPT. The distribution of RPTs resulting

from the application of an EO depends on the operator's parameters. In other words, an EO maps its parameters to an RPT.

We now define the decoding, fitness, and evolutionary operators, using the Merkle and Lamont [1] notation:

Since *H* is a nonempty set, let $f: \mathbb{R}^n \to \mathbb{R}$ be the objective function. If $D: H \to \mathbb{R}^n$ is total, i.e., the domain of *D* is all of *H*, then *D* is called a decoding function. *D* defines the feasible space of the solutions for the optimization problem. The mapping of *D* is not necessarily surjective; the range of *D* determines the subset of $\mathbb{R}^n$ available for exploration by the EA.

Define $T_s$, the fitness-scaling function, as $T_s : \mathbb{R} \to \mathbb{R}$. The fitness-scaling function converts the raw fitness scores returned by the fitness function (defined below) to values in a range suitable for the selection function. The selection function uses the scaled fitness values to select the parents of the next generation. The selection function assigns a higher probability of selection to individuals with higher-scaled values.

We define the fitness function as $\Phi \triangleq T_s \circ f \circ D$, where the objective function *f* is determined by the application and *D* and $T_s$ are design issues. Note that if the objective function is not known, the scaling function needs to be modified so it can solve an order problem versus the value problem that arises when the objective function is known.

To define EOs let $\mu \in \mathbb{Z}^+$ and $\mathrm{K}$ be the set of exogenous parameters. The mapping $X : \mathrm{K} \to \Upsilon\left(\Omega, \Upsilon\left[H^\mu, \bigcup_{\mu' \in \mathbb{Z}^+} H^{\mu'}\right]\right)$ is a generic EO. The set of EOs is denoted as $EVOP\left(H, \mu, \mathrm{K}, \Omega\right)$.

The three EOs—crossover, mutation, and selection—are roughly analogous to their similarly named counterparts in genetics. The application of them in EAs is strictly Darwin-like in nature, i.e., "survival of the fittest."

For the crossover operator let $C \in EVOP\left(H, \mu, \mathrm{K}, \Omega\right)$. If there exists $P \in H^\mu, \Upsilon \in \mathrm{K}$, and $s \in H$ such that one individual in the offspring population $C_\mathrm{K}(P)$ depends on more than one individual of *P,* then *C* is referred to as a crossover operator.

A mutation is defined in the following manner: let $U \in EVOP\left(H, \mu, \mathrm{K}, \Omega\right)$. If for every $P \in H^\mu, \Upsilon \in \mathrm{K}$, and $s \in H$ and if each individual in the offspring population $U_\mathrm{K}(P)$ depends on at most one individual of *P,* then *U* is called a mutation operator.

Finally, for a selection operator let $F \in EVOP(H, \mu, K \times \Upsilon(H, \mathbb{R})\Omega)$. If $P \in H^\mu, \Upsilon \in \mathrm{K}$, and $f : H \to \mathbb{R}$ in all cases and if *F* satisfies $s \in F_{\mathrm{K},\Phi}(P) \Rightarrow s \in P$, then *F* is a selection operator.

Please note that while we have used the Merkle and Lamont [1] notation so far, we will be switching to a certain extent to the Vose [2] and Wright and Vose [3] notation in the next section. We will, however, carry over the notations from above where needed, e.g., the search space *H* and *RPT*.

*1.2 Initial notation and ring theory*

We now introduce some EA notions from Wright and Vose [3]. Let our search space *H* consist of *d*-ary (modulo d) strings *s* of length *l*. Let $n = d^l$. Integers in the interval *[0, n)* are identified with elements of *H* through their *d*-ary representation. This correspondence allows them to be regarded as elements of the product group $H = d\mathbb{Z} \times ... \times d\mathbb{Z}$ (computed *l* times), where $d\mathbb{Z}$ denotes the integer's modulo *d*. The group operation on this product (addition modulo *d*) is denoted by $\oplus$, and the operation of component-wise multiplication (modulo *d*) is denoted by $\otimes$. Component-wise subtraction (modulo *d*) is denoted by $\ominus$, and $0 \ominus s$ is abbreviated as –*s*, with s defined as a column vector in $\mathbb{R}^l$, with its components being the *d*-ary digits. The notation $\bar{s}$ abbreviates $1 \ominus s$. The operation $\otimes$ takes precedence over $\oplus$ and $\ominus$, and all three bind more tightly than other operations except $s \to \bar{s}$, which is unary and has the highest precedence.

Given $s \in H$, let those *i* for which $s \otimes d^i > 0$ be $i_0 < i_1 ... < i_{m-1}$, where *m = #s*, and *#s* denotes the number of nonzero *d*-ary digits of *s*. The injection corresponding to *s* is the $l \times m$ matrix *S*, defined by $S_{i,j} = [i = i_j]$. To make explicit the dependence of *H* on string length, this dependence is abbreviated as $^l H$. The embedding corresponding to *s* is the image under *S* of $^m H$ and is denoted $H_s$ (we take the elements of $^m H$ and $^l H$ as column vectors). Integers in the interval (0; $d^m$) correspond to elements of *s* through *S*. Note that $H_s$ is an Abelian commutative group under the operation $\oplus$, and more generally it is a commutative ring with respect to $\oplus$ and $\otimes$.

An element $s \in H$ is called binary (even if *d > 2*), provided $s_i > 0 \Rightarrow s_i = 1$. The utility of embeddings follows from the fact that if *s* is binary, then each $i \in H$ has a unique representation $i = u \oplus v$, where $u \in H_s$ and $v \in H_{\bar{s}}$. This follows from the identity $i = i \otimes s \oplus i \otimes \bar{s} = u \oplus v$.

Next, we work with crossover, mutation, and what Wright and Vose [3] call the mixing scheme. We use primarily the work of Rowe *et al*. [4] to develop the necessary definitions.

To briefly describe the work in [4], it is supposed that the finite search space *H* has certain symmetries that can be described in terms of a group of permutations acting upon it. If crossover and mutation respect these symmetries, then these operators can be described in terms of a mixing matrix and a group of permutation matrices. [4] also examines the conditions under which certain subsets of *H* are invariant under crossover, which leads to a generalization of the term **schema**. [4] also notes that it is sometimes possible for the group acting on *H* to induce a group structure on *H* itself.

Let the search space *H* have the elements $u, v, \omega$, with $u, v$ being the parents and $\omega$ the child. Let *H* be acted upon by a finite group $(L, \circ)$, where *L* is a set of permutations of *H* that forms a group under function composition. That is, there is a mapping such that $L \times H \to H$. Let $\pi$ be a permutation of *H*. The action of $\pi \in L$ on some element $\omega \in H$ is denoted by $\pi(\omega)$. A transitive group action, denoted by $L(H)$, is reduced if it contains only the identity element. We assume $L(H)$ is reduced in the following. If it is not, then *L* is replaced by $L/H_L$.

The mixing scheme *M*, which describes the effect of crossover and mutation on a population, can be defined in the following manner:

> Let $\pi$ be the permutation matrix, with the *i, j*th entry given by $j - i = s$. Then, $(\pi_s u)_i = u_{i \oplus s}$. Denote the mixing scheme *M* by $M(u) = <...,(\pi_i u)^T M \pi_i u,...>$. The *i*th component function $RPT_i$ is the probability that *i* is the result of selection, mutation, and crossover. In vector form: $RPT = M \circ F$. This is slightly different from the Merkle and Lamont [1] formulation above. The difference, however, is due to the use of the intermediate product *M*, i.e., $M = C \circ U$, where *C* is crossover and *U* is mutation.

Now, let $a(u, v, \omega) = [b(u, v, \omega) + b(v, u, \omega)]/2$, where $b(u, v, \omega)$ is the probability that the crossover produces child $\omega$ from parents *u* and *v*. Crossover commutes with $L(H)$ iff $a[\pi(u), \pi(v), \pi(\omega)] = a(u, v, \omega)$ for all $u, v, \omega \in H$.

As for mutation, *U* also commutes with $L(H)$. The author leaves the reader to understand the necessary definitions and proof in Rowe *et al*. [4].

When mixing commutes with $L(H)$ the mixing scheme can be written as a mixing matrix *MM,* together with the associated set of permutation matrices $M(P)_\omega = p^T(\sigma_\pi MM \sigma_\pi^T)$, where $\pi \in L$ is chosen such that $\pi(0) = \omega$ and $MM_{u,v,} = a(u, v, 0)$ is the probability that parents *u* and *v* produce offspring $\omega$ after *C* and *U*.

If *H* has nontrivial subgroups $Q_0,...,Q_{t-1}$ such that for all $i, j \in \{0, 1, ..., l-1\}$ and $\omega \in H$:

1. $H = Q_0 \oplus,...,\oplus Q_{t-1}$
2. $i \neq j \Rightarrow Q_i \cap Q_j = \{0\}$
3. $\omega \oplus Q_i = Q_i \oplus \omega$

$H$ is the internal direct sum of the $Q_i$, which are normal subgroups of $H$. Each element $\omega \in H$ has a unique representation $\omega = \omega_0 \oplus, ..., \oplus \omega_{l-1}$, where $\omega_i \in Q_i$ and $l$ is the length of the string $s$. The map $\omega \to \langle \omega_0, ..., \omega_{l-1} \rangle$ is an isomorphism between $H$ and the product group $Q_0 \times, ..., \times Q_{l-1}$, where $\langle u_0, ..., u_{l-1} \rangle \oplus \langle v_0, ..., v_{l-1} \rangle = \langle u_0 \oplus v_0, ..., u_{l-1} \oplus v_{l-1} \rangle$ and $\odot \langle \omega_0, ..., \omega_{l-1} \rangle = \langle \odot \omega_0, ..., \odot \omega_{l-1} \rangle$. We note in passing that in [4] the concept of **schemata** (defined below) is generalized to capture this subgroup structure of the group action.

## 2. Group action, equivalence relations, and equivalence classes

As stated earlier, $H$ is acted upon by a finite group $(L, \circ)$, where $L$ is a set of permutations of $H$ (a nonempty set) that forms a group under function composition. For each $\pi \in L$ define the $n \times n$ permutation matrix as $(\sigma_\pi)_{u,v} = [u = \pi(v)]$ [b]. The set of all such permutation matrices forms a group under matrix multiplication that is isomorphic to $L$. Each permutation matrix can also be thought of as a linear map $\sigma_\pi : \mathbb{R}^n \to \mathbb{R}^n$.

It is easy to establish that when $H$ is acted upon by a finite group $(L, \circ)$ and when the $L$ subgroup represents all the valid permutations of $H$ (i.e., the automorphisms that arise because of the constraints of the optimization problem) we have an $L$-module.

We now define an orbit—a concept we will use to help establish equivalence relations and equivalence classes for EAs. An orbit of a point $\varsigma$ in $H$ is the set of elements of $H$ to which $\varsigma$ can be moved by the elements of $L$. We define the orbit of $H$ as $L(\varsigma) = \{\pi.\varsigma : \pi \in L\}$. If we denote the action of $L$ on $H$ by $L(H)$, we can define a stabilizer subgroup $H^L$ of $L$ as $H^L = \{\pi \in L : \forall u.\pi(\varsigma) = \varsigma\}$.

To show the connection between the group action and the partitioning of $H$ via equivalence relations we use Radcliffe [5]. Let the search space $H$ be taken to be the collection of equivalent classes, where $H$ is defined as the space of phenotypes, i.e., the elements $\varsigma$ that are subsets that partition $H$ so that $H/\sim$ exists ($\sim$ is an equivalence relation on $H$).

Because of the defining properties of a group, a set of orbits (points $\varsigma$ in) of $H$ under the action of $L$ form a partition of $H$. The associated equivalence relation for this definition of partition is $\varsigma \sim \ell$ iff there exists a $\pi$ in *L,* where $\pi.\varsigma = \ell$. This relationship has been shown to be true above. The orbits are, therefore, the equivalence classes under this relation, and the two elements $\varsigma$ and $\ell$ are equivalent iff their orbits are the same, i.e., $L\varsigma \sim L\ell$.

---

[b] The set of $n \times n$ invertible matrices with real (or complex) coefficients is a group under matrix multiplication, with its identity element being the identity matrix $I_n$. This group is called the general linear group and is usually denoted as $GL(n, \mathbb{R})$ (or $GL(n, \mathbb{C})$).

To connect equivalence relations and classes to the EA terms schemata and schema, respectively, we note that schemata are members of a schemata family. Any $C \in H$, where *C* is a chromosome, represents a schemata family per Radcliffe [5] and Holland [6]. The elements of the schemata family (chromosomes) are subsets that partition $H$. Thus, a schema is an equivalence class, and a schemata is an equivalence relation.

In Rowe *et al*. [4] a similar definition of schema as orbits/equivalence classes is arrived at by using subgroups. In brief, given an *N* that is a set of subgroups of *L* and an *A* that is a normal subgroup of *L*, [4] develops a definition where a schema is defined as the orbits of *A.*

The set of all orbits of *H* under the action of *L* is *H/L*; *H/L* is called the quotient of the action. In geometric situations *H/L* is called the orbit space, while in algebraic situations *H/L* is the space of coinvariants and is written as $H_L$. This is in contrast to invariants (fixed points), which are denoted as $H^L$. The author notes that the various invariant properties discussed in Vose [2] and Rowe *et al*. [4] can be derived using $H^{L\,c}$.

### 3. Optimization: covering set and sufficiency

Many papers, starting with Holland [6], explain the computational behavior of EAs by arguing that EAs compare the equivalence relations of the search space and then allocate more trials to the equivalence relation of the search space that has the highest fitness. These equivalence relations, as stated before, are the schemata $\varsigma \sim \ell$. EAs can therefore be seen to stochastically "hill climb" in the space of equivalence relations rather than hill climb in the space of *d*-strings (see Salomon and Arnold [7] for a detailed discussion). And, as with other hill-climbing strategies, an EA will perform poorly if there is an absence of information or if that information is misleading (deceptive). Therefore, for the hill climbing to be successful a set of equivalence relations must induce a useful representation of the search space.

To establish the existence of this useful representation, we note (following [5]) that in linear algebra a basis is a set of linearly independent vectors, which in a linear combination represents every vector in the given vector space. A similar basis can be established for EAs (proof to follow) if for the equivalence relations

---

$^c$ See in particular Vose [2].

that partition *H*, there is a basis $\Im$ for the set of equivalence relations *ER*, i.e., a general equivalence relation can be decomposed as an intersection of basic equivalence relations in $\Im$. We define two equivalence relation properties to use in our proof that the necessary coverage of *H* can be generated:

Let $E(H)$ be the set of all equivalence relations in *H*, and let it be understood that the single equivalence relation $\varepsilon \in E(H)$ exists. We define a partial representation function using the equivalence class induced by $\varepsilon$ as $\rho_\varepsilon : H \to \Xi_\varepsilon$ by $\rho_\varepsilon(x) \triangleq [x]_\varepsilon$, where $[x]_\varepsilon$ is the equivalence class of *x* under $\varepsilon$, i.e., $[x]_\varepsilon \triangleq \{y \in H : \varepsilon(x, y,) = 1\}$.

Given any $\Im \subset E(H)$ where $\Im = \{\varepsilon_1, ..., \varepsilon_n\}$, we define an evolutionary representation function $\rho_\Im \triangleq (\rho_{\varepsilon_1}, ..., \rho_{\varepsilon_n})$, such that $\rho_\Im : H \to \Xi_\Im$ with $\rho_\Im(x) = ([x]_{\varepsilon_1}, ..., [x]_{\varepsilon_n})$. The function $\rho_\Im$ maps each solution in *H* to the vector of basic equivalence classes/schema/orbits to which it belongs.

Given a basis $\Im$ for a set of equivalence relations $ER \subset E(H)$, define *C*—the space of chromosomes—to be the image of *H* under $\rho_{ER}$:
$C \triangleq \underset{\rho_{ER}}{\operatorname{Im}} \equiv \rho_{ER}(H) \equiv \{\xi \in \Xi_{ER} : \exists x \in H : \rho_{ER}(x) = \xi\} \subset \Xi_{ER}$, where $\xi$ is an equivalence class.

To define coverage let $x, y \in H$, $\varepsilon$ be an equivalence relation, and *ER* the set of equivalence relations. Coverage exists if $\forall x \in H$ and $\forall y \in H$, $\{x\} \exists \varepsilon \in ER : \varepsilon(x, y) = 0$.

We define the basis of a covering set by letting $\Im$ be a basis for a set of equivalence relations, i.e., $ER \subset E(H)$, that covers *H*. Therefore, $\Im$ covers *H*. The proof of this statement follows immediately: since every equivalence relation $\varepsilon$ in *ER* can be expressed as an intersection of some members of the basis $\Im$, then if $\varepsilon$ distinguishes between two solutions, at least one of the equivalence relations in $\Im$ can be decomposed.

As noted above, coverage is important because, if a set of equivalence relations covers *H*, then specifying to which equivalence class a solution belongs for each of the equivalence relations in the set is sufficient to identify a solution uniquely.

## 4. Dynamical systems and group actions

We start this section with a brief summary of Vose's [2] work on EAs and dynamical systems:

Vose's EA model can be characterized as a discrete dynamical system and a kind of map. Iterating the map simulates the trajectory of the EA, where the next population vector becomes the input to the next generation of the EA. This forms a sequence of population vectors $P_1, ..., P_n$. This sequence is the trajectory of the EA model through the population space. We make extensive use of these trajectories as we describe how EAs solve global optimization problems.

We now develop the necessary connections between group actions and dynamical systems before proceeding to solve global optimization problems.

A linear transformation is one of the simplest ways to describe the relationship between two vector spaces. Over linear subspaces with a countable basis, linear transformations can be represented by matrices. It is often desirable to represent a linear transformation as being as characteristic as possible; this leads to the notion of identifying a matrix by its canonical form. The canonical form is most frequently expressed in terms of matrix decomposition.

Many types of canonical forms exist in the literature. Those feasible for numerical computation include the spectral decomposition of symmetric matrices, the singular value decomposition for rectangular matrices, and the Schur decomposition for general square matrices. We mention these canonical forms, since we will refer to Schmitt's [8] [9] important work on EAs and his use of spectral decomposition in a later section of this paper.

Most matrix decompositions are calculated through iterative procedures, their success being evidenced by the many discrete methods that are available. Our goal in this section is to recast some of these iterative schemes as dynamical systems via group actions. We do this to stay with the extensive use we've made of group action in understanding EAs and to show that this characterization (the use of group, action, and orbit) is a generalization of many of the methods that rely on matrix decomposition to understand EAs.

Before we start our formal development we need to know that the meaning of a canonical form in this paper is understood in a much broader context than just matrix factorizations.

We point out that—as a whole—the procedure for finding the simplest form in most applications is to follow the orbit of certain matrix group actions on the underlying matrix. This connection should not come as a surprise; the representation of a group by its homomorphisms into bijective linear maps over a certain vector space is a well-known concept. For groups whose elements depend on continuously varying parameters, so do the corresponding matrix representations. The obvious advantage of this relationship is that we have the group structure on one side and the matrix structure on the other side.

The question to ask now becomes, in what canonical form can a matrix or a family of matrices be linked to the orbit of a group action? The choice of the group, the definition of the action, and the targets the action is intended to reach will effectuate the various paths of transitions and thus the algorithms.

With group, action, and orbit for EAs in place (see Section 2 above), we now need a properly defined dynamical system—either continuous or discrete—that has integral curves or iterates that stay on the specified orbit and connect one state to the next. We develop a general framework of the **projected gradient approach** to help construct useful dynamical systems. The projected gradient flows from continuous group actions are easy to formulate and analyze. And, sometimes they are able to tackle problems that are seemingly impossible to resolve by conventional discrete methods.

As a side note: gradient flow, where finding its stable equilibrium point is the goal, has been developed thoroughly via group, action, and orbit. The reader is referred to the work of Tam [10].

The idea of projected gradient flows stems from the constrained least-squares approximation to a desirable canonical form. Using general notation, from a given matrix *A* in a subset *V* of matrices of fixed sizes, the constraint on the variable is that the transformation of *A* must be limited to the orbit *OrbG(A)* (in Section 2 we defined the orbit of *H* as $L(\varsigma) = \{\pi.\varsigma : \pi \in L\}$ ). The orbit is determined by a prescribed continuous matrix group *G* and a group action $\mu : G \times V \to V$ (this mapping using *L* and *H* is used in [4]). The objective function itself is built with two additional limitations. One is a differentiable map $f : V \to V$ designed to regulate certain inherent properties such as symmetry, diagonal, isospectrality, low rank, or other algebraic conditions ([3] and [4] as well as [8] [9] use some of these properties in their discussion of group properties and spectral theory). The other is a projection map $P : V \to P$, where *P* denotes the subset of matrices in *V* that carries a certain desirable structure, i.e., the canonical form. The set *P* could be a singleton, an affine subspace, a cone, or another geometric entity.

Consider the functional $F : G \to \mathbb{R}$, where 
$F(Q) := \frac{1}{2} \| f(\mu(Q, A)) - P(\mu(Q, Q)) \|_F^2$. The goal is to minimize *F* over the group *G*. The meaning of this constrained minimization is that, while staying in the orbit of *A* under the action of *μ* and maintaining the inherent property guaranteed by the function *f*, we look for the element $Q \in G$, such that the matrix $f(\mu(Q, A))$ best realizes the desired canonical structure in the sense of least squares.

The projected gradient flow approach can be formulated as a dynamical system $\frac{dQ}{dt} = -\text{Proj}_{T_Q G} \nabla F(Q)$, where $T_Q G$ and $\nabla F(Q)$ stand for the tangent space of the group *G* and the gradient of the objective functional *F* at *Q*, respectively.

One advantage of working with a matrix group is that its tangent space at every element *g* has the same structure as $\ell = T_e G$ at the identity element *e* of *G*. More specifically, the tangent space of any element *Q* in *G* is a translation of $\ell$ via the relationship $T_Q G = Q\ell$. Thus, the projection $-\text{Proj}_{T_Q G} \nabla F(Q)$ is fairly easy to do, once the tangent space $\ell$ is identified.

We have now demonstrated how group actions can serve as the fundamental coordinate transformations that lead to canonical forms. It comes as no surprise—but rather as a necessity—that many of the dynamical systems and numerical algorithms originally developed over the Euclidean space need to be redeveloped over manifolds. We discuss this in the next section when we work with the methods of Absil *et al*. [11].

**5. Global optimization via dynamical systems**

In the following discussion we show that different dynamical systems are in play for the unconstrained case and the constrained case. To start our examination of EAs using dynamical systems theory, consider the gradient dynamical system $\dot{x} = -\nabla f$, where $\arg\min_{x \in H} f$ is taken as an unconstrained nonlinear programming problem. Chiang *et al*. [12] show that each local optimal solution of $\arg\min_{x \in H} f$ corresponds to a stable equilibrium point of $\dot{x} = -\nabla f$, where $\dot{x} = -\nabla f$ is a special class of the general nonlinear dynamical system $\dot{x}(t) = b[x(t)]$ and where the state vector $x(t)$ belonging to the Euclidean space $\mathfrak{R}^n$ and $f : \mathfrak{R}^n \to \mathfrak{R}^n$ satisfies the sufficiency condition for the existence of unique solutions.

[12] also shows that if $x_s$ is a stable equilibrium point, then it is a local optimal solution for $\arg\min_{x \in H} f$. And if $x_s$ is a local solution of $\arg\min_{x \in H} f$, then it is a stable equilibrium point of $\dot{x} = -\nabla f$. [12] proves that in the unconstrained nonlinear case the search space *H* is the union of the closure of the regions/basins of attraction of all the stable equilibrium points. [12] also shows that two adjacent regions are separated by the intersection of their stability boundaries (called a "joint stability boundary" in [12]). And, to identify an adjacent stable equilibrium point from the current stable equilibrium point, a mechanism (be it deterministic or stochastic) needs to move across the intersection from one stability region to its adjacent stability region.

If we assume the first set of equivalence relations generated by an EA places the equivalence relation at or near a stable equilibrium point, what needs to happen next—after the expected temporary stasis that results from its being at or near a fixed point—is for the EA to move toward its stability boundary. A stability boundary is the boundary of a stability region. It can be defined as being contained in the union of the stable manifolds of the equilibrium points $x_s$ on the stability boundary. This definition assumes all the equilibrium points $x_s$ of the gradient system are hyperbolic, which Vose [2] has shown to be true.

If the current fixed point is not part of a basin of attraction, the "escape" of the EA from its current area via the fitness evaluations of successive equivalence relations is almost certain. This is because the dynamics near an unstable fixed point mean the future equivalence relation derived from the local dynamics will include at some point the related nontrivial unstable space. This subspace's spectrum is exterior to the unit disk, and its unstable manifold converges with the next stable fixed point $x_s^n$ (see [12], Theorem 3.4).

The "escape" from a basin of attraction (stability region) is more difficult, since the unstable space in this case is trivial. For the EA to continue its exploration of *H* it will more than likely make a slower "climb" as it looks for an exit point. As the equivalence relations change the value of $\arg\min_{x \in H} f$ will increase, assuming the local optimal solution is where the EA started. The values of the objective function will increase until they reach a point where the value of $\arg\min_{x \in H} f$ decreases. This point will be seen as an "exit" point by the EA; this exit point equates to the intersection of the stability boundary and the curve connecting the sequence of values (see both [12] and [2] concerning the importance of the Lyapunov function in this sequence of events). The next fixed point $x_s^n$ has now been found, and the search space *H* will continue to be examined.

If $\arg\min_{x \in H} f$ is a constrained nonlinear programming problem, the task of finding and evaluating each of the feasible regions derived via the decoding function *D* is very difficult. This difficulty arises because the set of feasible regions is usually nonconvex and disconnected, i.e., it is composed of several disjoint-connected feasible regions.

It can be shown that the set of feasible regions is a smooth manifold (see for example, Absil *et al*. [11]) and that each feasible region corresponds to a stable equilibrium manifold of the nonlinear system (a stable equilibrium manifold is a generalized concept of a stable equilibrium point). In particular, if the set is a feasible region, then the set is a stable equilibrium manifold of a quotient gradient system whose vector field is the constraint set *H(x)* (see Lee [13]). So, as with the unconstrained nonlinear problem, the EA in the constrained nonlinear case has a dynamical system—in this case a quotient gradient system.

To start our analysis we state the quotient gradient system associated with the constraint set characterized by *H(x)* as being $\dot{x} = -JH(x)^T H(x)$, with *JH(x)* being the Jacobian matrix of the vector *H(x)*. Since *H* is composed of several disjoint-connected feasible regions each of which the EA needs to explore, we develop a projected gradient dynamical system where all the local optimal solutions correspond to all the stable equilibrium points of the nonlinear dynamical system.

Our projected gradient system is $\dot{x} = -P_H[x(t)]\nabla f[x(t)]$, where $x(0) = x_0 \in M$ and *M* is the manifold. The projection matrix is
$P_H(x) = [I - JH(x)^T (JH(x)JH(x))^T DH(x)] \in \mathbb{R}^{n \times n}$, where *I* is the count related to the constraints. For example: $h_i > 0$, where $i \in I = \{1,...,l\}$.

The related positive semidefinite matrix for every $x \in M$ is
$P_H(x) = [I - JH(x)^T (JH(x)JH(x))^T DH(x)] \in \mathbb{R}^{n \times n}$. $P_H(x)\nabla f(x)$ is the orthogonal projection of $\nabla f(x)$ to the tangent space $T_x M$, which means $P_H(x)\nabla f(x) \in T_x M$ for all $x \in M$. Note that every trajectory of $\dot{x} = -P_H[x(t)]\nabla f[x(t)]$, starting from $x_0 \in M_k$, stays in $M_k$. Therefore, $M_k$ is an invariant set, i.e., a set of fixed points, of $\dot{x} = -P_H[x(t)]\nabla f[x(t)]$.

Now that we have established a correspondence between all the local optima and all the stable equilibrium points of the nonlinear dynamical system, we show how the EA moves from the current local optimal solution and approaches another local optimal solution.

We restate this problem as one of how to escape from the region of stability of the corresponding stable equilibrium point. Using the projected gradient system, we note that escaping from a region of stability of the corresponding stable equilibrium point is $\dot{x} = -P_H[x(t)]\nabla f[x(t)]$ and entering the region of another stable equilibrium point is $\dot{x} = -P_H[x(t)]\nabla f[x(t)]$. Once a system trajectory lies inside the stability region of a stable equilibrium point, the ensuing system trajectory converges to the stability equilibrium point, which is a local optimal solution of the constrained nonlinear optimization problem.

Finally, to ensure that most if not all the feasible region is explored by the EA, we remember that the EA near a stable fixed point/region experiences a temporary stasis. This temporary stasis results in an expanded search of the feasible region. The length of time the EA explores the current feasible region is driven primarily by its design parameters, e.g., mutation, crossover, string length, and fitness function. So, although it can be stated that an EA has the ability to fully explore a feasible region, this ability is ultimately a potential. An individual EA's design parameters determine how successful the search of the feasible region is.

## 6. EAs and spectral theory

Schmitt [8] [9] uses spectral theory and the properties of time-inhomogenous Markov chains to understand the workings of EAs as well as how EA design parameters can be set. He is very successful in terms of giving guidance on what to consider when using crossover, mutation, fitness functions, and—most importantly—population size.

To quote from [9]:

> What makes our approach different is that we do not attempt to unite the genetic operators crossover, mutation and selection into one operator which is subsequently analyzed. We rather analyze the genetic operators separately to isolate key properties: 1) crossover plays a dual role enhancing mutation in the mixing phase of the algorithm as well as enhancing selection in some cases, 2) mutation is responsible for weak ergodicity and the flow away from uniform populations, 3) fitness selection is responsible for contraction towards uniform populations, 4) mutation-selection is responsible for convergence to uniform populations in the zero mutation rate limit, and 5) all three genetic operators act together to obtain the steady-state flow inequality which shows convergence to global optima.

The main results of Schmitt's using (1)–(5) are: (1) a general-purpose, scaled, converging genetic algorithm is developed whose setup is quite similar to that of the simulated annealing algorithm; (2) explicit cooling schedules for not-necessarily commuting mutation-crossover (such as the gene-lottery crossover)and exponentiation schedules for fitness selection exist; and (3) no conditions are attached, i.e., the fitness function needs not be injective, and the population size can stay small and controllable. Clearly, Schmitt's work on the need to examine the evolutionary operations (EVOPs) separately and also the need to demonstrate that the necessary convergence is present is a major step forward in effectively generating the design parameters that will help solve the gradient systems discussed above.

As stated in Section 4, our paper has stayed within the group action context. Our use of group, action, and orbit as well as our broader development of canonical form cause us to differ from Schmitt. Here we briefly develop the needed group theory to show that spectral theory is aligned with or very nicely drops out of the work of Wright and Vose [3], Rowe *et al*. [4], and others.

One could begin with generalized eigenfunctions to start the spectral theory development, but it is simpler to construct a group algebra, the spectrum of which captures the Fourier transform's basic properties. This is carried out by means of Pontryagin duality. With Pontryagin duality, complex-valued functions on a finite Abelian group have discrete Fourier transforms that are functions on the dual group, which is a (noncanonical) isomorphic group. Moreover, any function on a finite group can be recovered from its discrete Fourier transform. We note at this point the importance of Fourier transforms and Abelian groups in [3] and [4]. The Pontryagin dual of a topological Abelian group *A* is locally compact if and only if *A* is locally compact.

A spectrum is a bounded operator that is a generalization of the concept of eigenvalues for matrices. Specifically, a complex number $\lambda$ is said to be in the spectrum of a bounded linear operator $T$ if $\lambda I - T$ is not invertible, where $I$ is the identity operator. The study of spectra and related properties is known as spectral theory.

A topological Abelian group (TAG) is a topological group that is also an Abelian group. That is, a TAG is both a group and a topological space where the group operations are continuous and the group's binary operation is commutative (the importance of continuous group operations and binary commutativity is discussed in an earlier section of this paper and also in [3] and [4]). The theory of topological groups also applies to TAGs.

To connect spectrum, Pontryagin duality, and topological spaces, we note that the theory of unitary representations of groups is closely connected with harmonic analysis. In the case of an Abelian group *G*, a fairly complete picture of the representation theory of *G* is given by Pontryagin duality. In general, the unitary equivalence classes of irreducible unitary representations of *G* make up its unitary dual. This set can be identified with the spectrum of the *C\**-algebra (a group algebra) associated to *G* by the group *C\**-algebra construction. This is a topological space.

Schmitt [8] makes use of spectral radius and spectral calculus to estimate such things as the contraction/mixing properties of the combined crossover-mutation operator in an EA. Since the spectral radius of a finite graph is defined to be the spectral radius of its adjacency matrix, Schmitt is therefore starting to solve the EA graph problem with the use of spectral radii. He does not directly solve the related graph problem (much too difficult at the time), but his use of tools such as Frobenius and the Spectral Mapping Theorem can now be applied directly to finite (EA) graphs and to their stochastic matrices as shown by the work of Escola [14] on spectral graph theory. This is not to take away from any of Schmitt's very important work on EAs but rather to note that what was an intractable graph problem a decade or more ago is now tractable.

The spectral radius can be generalized to the **joint spectral radius** when there are sets of stochastic matrices—a form Schmitt [8] uses. The joint spectral radius was introduced for its interpretation as a stability condition for discrete-time dynamical systems (such as simple EAs). There is an intimate connection between the joint spectral radius and the Lyapunov exponent on path-complete graphs[d]. This last point connects nicely with Schmitt's statement on graphs and

---

[d] See A.A. Ahmadi, R. M. Jungers, P. Parrilo, M. Roozbehani, "Analysis of the joint spectral radius via Lyapunov functions on path-complete graphs." Proc. of HSCC '11, Chicago, 2011.

EAs in [9] as well as with Vose's comments [2] concerning the importance of the Lyapunov function in determining the "exit" points for EAs.

**7. Optimization: two open problems**

In this section we address two open EA problems: (1) global optimization when the dynamical system is chaotic and (2) global optimization when the optimization surface is fractal.

In the chaotic case (which Vose [2] touches on briefly) the issue facing the EA is the piling up of unstable manifolds via their equivalence classes. The existence of these foliations explains in part why EAs have problems escaping from manifolds/fixed points: the foliations generate an attracting set of their own.

To elaborate, the complicated piling up of the unstable manifolds on themselves suggests that the invariants may have very rough densities in the direction of the transversal to the foliations of the unstable manifolds. Therefore, the space of equivalence classes does not exist in general (as a measurable space) because of the folding and accumulation of unstable manifolds. Instead, we can work with local unstable manifolds that allow (transform) the roughness of the invariants in such a way that the density functions become differentiable. This step—in the author's view—is a closed door to EAs in terms of achieving a solution. Please see Eckmann and Ruelle [15] for an excellent treatment of this problem.

In regard to fractal surfaces Back [16] has shown the difficulties EAs have in solving problems with a fractal dimension. The main reason for the EAs' relative failure is the underlying homology of the map used.

As shown by a number of authors, applying the algebra used above does not solve the problem. It only provides a starting framework. And, as far as this author knows, a general approach does not exist for the problem; instead, one must look at the specific fractal in greater detail.

Now, we need not be so abstract. We could proceed recursively on successive graph approximations to the fractal—supposing such graphs exist and are suitably well-behaved—by considering more or less discrete de Rham cohomology on the approximating graphs. This is not unlike the simplicial cohomology of a simplicial complex. Perhaps there is a limiting procedure here not unlike that which prevails in Cech cohomology that could produce the final de Rham cohomology of the fractal itself.

**8. Summary and conclusions**

In this paper we make extensive use of group theory and dynamical systems to explain how EAs solve global optimization problems. We start with a basic statement about EAs (without constraints) and then use the work of Merkel and Lamont [1] to establish a baseline random heuristic for EAs.

We use the work of Wright and Vose [3] and Rowe *et al*. [4] to establish the importance of group and ring theory in understanding EAs and to link that to the group action of $(L, \circ)$ on *H*. From the EAs' group action we are able to establish orbits and invariants that lay the groundwork for equivalence classes and equivalence relations. We link equivalence classes and relations to schema and schemata, respectively.

We show, using the work of Radcliffe [5], that sufficient coverage via the use of equivalence classes and relations generates a unique solution for an optimization problem. Using group, action, and orbit, we develop a canonical matrix form. We then develop a properly defined dynamical system, either continuous or discrete, such that its integral curves or iterates stay on the specified orbit and connect one state to the next. We develop a general framework of the projected gradient approach to help construct useful dynamical systems and to show how EAs solve both constrained and unconstrained global optimization problems. We also discuss the important work of Schmitt [8] [9] in setting design parameters.

Finally, we discuss two open problems in EA optimization: chaotic dynamical systems and fractal surfaces. We show that optimization problems that are chaotic dynamic systems are difficult for an EA to solve because of the foliation of unstable manifolds. However, optimization problems with fractal surfaces may have potential solutions, if one works with the individual fractal surfaces.

In closing, we briefly examine design issues, a subject we touched on when discussing the work of Schmitt [8] [9]. We note there is no guarantee an EA will move to the space of greater stability, which is needed if a global optimization problem is to be solved. The next stable fixed point $x_s^n$ is just that, the next stable fixed point. The number of generations needed to get to the fixed point of greater stability may be too few in terms of the number of generations remaining in the EA. It may also be that the current mutation and/or crossover parameters may not allow the EA to get any closer than it is now. Or, it could be the length of the string *s* may not be sufficient to get the EA close to the global minimum. All these problems are addressed very effectively in Schmitt [8] [9], so practitioners would do well to familiarize themselves with his work.

In mentioning design issues we emphasize our view that exploratory data analysis (EDA) needs to be done for each EA problem, since EDA can help set the initial estimates of the EA's parameters. It does so by revealing the geometric and topological nature of the problem via the use of such tools as isomaps and computational algebra programs, along with the computational topology tools that are just becoming available. Understanding the geometry and topology of the problem (as seen by the importance of the dynamical systems analysis done above) can be helpful in determining whether, for example, all the constraints are better modeled at once or whether the constraints should be broken into two (or more) groups. In a separate paper, we have determined, using isomaps and computational algebra, that for some EA problems a two-stage EA is needed, i.e., a limited set of constraints is modeled in the first EA, while the remaining constraints are modeled in a second EA that takes the first EA's output as part of its input. Our use of isomaps and algebra to determine the geometry of the system does not uncover its topology, however. The work of Carlsson and others (see Carlsson [17] for a very good introduction to computational topology) shows the effectiveness of using topological tools and in particular their ability to preserve the underlying topology **and** geometry of the data cloud. These insights can be very valuable when we are using EAs.

**Vitae**

Andrew Clark is the Manager of Alternative Investment Research at Lipper. Andrew's most recent articles have appeared in the *Journal of Indexes*, the *Journal of Index Investing*, the *Journal of Investing*, and *Physica A*. His shorter pieces have appeared on websites such as IndexUniverse.eu, Seeking Alpha, Mine Web, Hedge Week, and Hubbis. In 2013 he will have a chapter on commodities appear in the book *Alternative Investments–Balancing Opportunity and Risk* (to be published by John Wiley). His recent work on evolutionary algorithms and optimization appeared as a chapter in the LNCS series by Springer in 2012.